%% file: oursland2024_nn_use_dm.tex
\documentclass[11pt]{article}

\usepackage{amsmath, amssymb}
\usepackage{graphicx}
\usepackage{natbib}
\usepackage{geometry}
\usepackage{hyperref}
\usepackage{booktabs}
\usepackage{authblk}
\usepackage{caption}
\usepackage{float}
\usepackage{subcaption}  

\title{Neural Networks Use Distance Metrics}
\author{Alan Oursland}
\affil{\textit{alan.oursland@gmail.com}}
\date{November 2024}

\begin{document}

\maketitle

\begin{abstract}
\input{abstract}
\end{abstract}

\input{introduction}
\input{related_work}
\input{background}
\input{design}
\input{results}
\input{discussion}
\input{conclusion}

\bibliographystyle{plainnat}
\bibliography{references}

\appendix

\input{appendix_statistics}

\end{document}

%% file: abstract.tex
We present empirical evidence that neural networks with ReLU and Absolute Value activations learn distance-based representations. We independently manipulate both distance and intensity properties of internal activations in trained models, finding that both architectures are highly sensitive to small distance-based perturbations while maintaining robust performance under large intensity-based perturbations. These findings challenge the prevailing intensity-based interpretation of neural network activations and offer new insights into their learning and decision-making processes.

%% file: introduction.tex
\section{Introduction}

The foundation for interpreting neural network activations as indicators of feature strength can be traced back to the pioneering work of McCulloch and Pitts in 1943 \citep{mcculloch1943logical}, who introduced the concept of artificial neurons with a threshold for activation. \footnote{The implementation for this work can be found at \url{https://github.com/alanoursland/neural_networks_use_distance_metrics}.} This concept, where larger outputs signify stronger representations, was further developed by Rosenblatt's 1958 perceptron model \citep{rosenblatt1958perceptron} and has persisted through the evolution of neural networks and deep learning \citep{schmidhuber2015deep}. Throughout this evolution, the field has largely upheld this interpretation that larger activation values indicate stronger feature presence -- what we term an \emph{intensity metric}. However, despite the remarkable success achieved through this lens, the statistical principles underlying neural network feature learning remain incompletely understood \citep{lipton2018mythos}.

This work builds on our recent theoretical framework \citep{oursland2024interpreting} that proposed neural networks might naturally learn to compute statistical \emph{distance metrics}, specifically the Mahalanobis distance \citep{mahalanobis1936generalized}. Our analysis suggested that smaller node activations, rather than larger ones, might correspond to stronger feature representations. While this previous work established a mathematical relationship between neural network linear layers and the Mahalanobis distance, we need empirical evidence to determine whether networks actually employ these distance-based representations in practice.

We use systematic perturbation analysis \citep{szegedy2013intriguing, goodfellow2014explaining} to provide empirical evidence supporting the distance metric theory proposed in our previous work. Using the MNIST dataset \citep{lecun1998gradient}, we modify trained models by independently manipulating distance and intensity properties of network activations. By analyzing how these perturbations affect model performance, we identify which properties -- distance or intensity -- drive network behavior. Our investigation focuses on two key questions:

\begin{itemize}
    \item Do neural networks naturally learn to measure distances rather than intensities when processing data distributions?
    \item How do different activation functions (ReLU and Absolute Value) affect the type of statistical measures learned by the network?
\end{itemize}

Our results show that networks with both ReLU and Absolute Value activations are highly sensitive to distance-based perturbations while maintaining robust performance under intensity perturbations, supporting the hypothesis that they utilize distance-based metrics. These findings not only validate our theoretical framework but also suggest new approaches for understanding and improving neural network architectures.

%% file: related_work.tex
\section{Prior Work}

In 1943 McCulloch and Pitt introduced a computation model of a neuron to explore logical equations in biological brains \citep{mcculloch1943logical}. Their definition $\text{TRUE} = (Wx > b)$ marks the beginning of our path using intensity metrics. Rosenblatt adapted this into an activation value $y = f(Wx + b)$ in 1957 with the perceptron, further solidifying the intensity metric interpretation \citep{rosenblatt1957perceptron}.

The development of multilayer perceptrons (MLPs) and the backpropagation algorithm enabled the training of deeper networks with continuous activation functions. \citep{rumelhart1986learning,lecun1989backpropagation,hornik1989multilayer} The interpretation of activations continued to focus on larger values as being more salient, reflected in visualizations of activations and analyses of feature maps, where stronger activations were highlighted. \citep{zeiler2014visualizing,yosinski2015understanding,olah2017feature,erhan2009visualizing}

The rise of deep learning, with the widespread adoption of ReLU and its variants, further reinforced the intensity metric interpretation by emphasizing the importance of large, positive activations. \citep{nair2010rectified,glorot2011deep} Visualization techniques, such as saliency maps and Class Activation Mapping (CAM), often focused on highlighting regions with high activations. \citep{simonyan2013deep,zhou2016learning} Similarly, attention mechanisms, which assign weights to different parts of the input, often rely on the magnitude of these weights as indicators of importance. \citep{bahdanau2014neural,vaswani2017attention}

While the intensity metric interpretation has been dominant, recent work has highlighted its limitations. \citep{rudin2019stop} Considering the relationships between activations, particularly through distance metrics, offers a promising avenue for understanding neural network representations. \citep{goodfellow2014explaining,madry2017towards,szegedy2013intriguing} Distance-based methods, such as Radial Basis Function (RBF) networks and Siamese networks, demonstrate the potential of incorporating distance computations into neural network architectures and interpretation. \citep{broomhead1988radial,bromley1994signature,schroff2015facenet} This approach could lead to more nuanced and effective representations.

%% file: background.tex
\section{Background}

In our previous work, \textit{Interpreting Neural Networks through Mahalanobis Distance}, we established a mathematical link between linear nodes with absolute value activation functions and statistical distance metrics. \citep{oursland2024interpreting} This framework suggests that neural networks may naturally learn to measure distances rather than intensities.

We explore this idea within the MNIST dataset, a well-known digit recognition problem that offers a structured environment for examining neural network behavior \citep{lecun1998gradient}. MNIST's clear feature structure and abundant prior research make it ideal for investigating core properties of neural network learning.
A \emph{distance metric} quantifies how far an input is from a learned statistical property of the data \citep{deza2009encyclopedia}. While an \emph{intensity metric} reflects a confidence level — larger values indicate higher certainty that the input belongs to the node's feature set. This dual interpretation of a node's output — either as a measure of distance or as confidence in feature presence — can help us understand the nature of neural network learning. For instance, an intensity filter could be viewed as a \emph{disjunctive distance metric} that measures how close an input is to everything the target feature is not.

\subsection{From Theory to Practice}

The distinction between distance and intensity metrics becomes critical when analyzing network behavior. Traditional interpretations suggest that the node detects the \emph{presence of 0}, with higher activation values indicating greater confidence. However, another possibility is that the node measures the \emph{distance from class embeddings that are not 0} — that is, how different the input is from all other digits. Both interpretations lead to the same result, but the disjunctive distance-based view aligns more closely with known statistical distance metrics. While there may be statistical intensity metrics, we have yet to identify one that models confidence signals in the same way.
This reframing suggests that neural networks might fundamentally operate by \emph{comparing inputs to learned prototypes} rather than \emph{assessing the strength of individual features}. However, proving this requires more than mathematical relationships; we need empirical evidence that networks indeed learn and use distance metrics in practice. This leads to several key questions:

\begin{itemize}
    \item Do neural networks naturally learn to measure distances rather than intensities?
    \item How can we experimentally distinguish between distance-based and intensity-based feature learning?
    \item What evidence would convincingly demonstrate which interpretation better reflects network operation?
\end{itemize}

These questions inform our experimental design, which uses controlled perturbations to test the nature of the learned features. By independently manipulating the distance and intensity properties of network activations, we can determine which aspects truly drive network behavior.

Our investigation focuses not on proving specific mathematical relationships but on demonstrating that \emph{distance-based properties}, rather than \emph{intensity-based properties}, govern network performance. This approach aims to improve our understanding of how neural networks process information and may lead to more effective network design and analysis methods \citep{montavon2018methods,samek2019explainable}.

%% file: design.tex
\section{Experimental Design}

To empirically investigate whether neural networks naturally learn distance-based features, we designed systematic perturbation experiments to differentiate between distance-based and intensity-based feature learning. This experimental framework directly compares these two interpretations by examining how learned features respond to specific modifications of their activation patterns. We hypothesize that perturbing the "true representation" will result in a drop in model accuracy. 

We train a basic feedforward model on the MNIST dataset to test our hypotheses. Our goal is to obtain a robust model for perturbation analysis, not to optimize model accuracy. The network processes MNIST digits through the following layers:

\begin{equation}
    x \rightarrow \text{Linear}(784) \rightarrow \text{Perturbation} \rightarrow \text{Activation (ReLU/Abs)} \rightarrow \text{Linear}(10) \rightarrow y
\end{equation}
 
The perturbation layer is a custom module designed to control activation patterns using three fixed parameters: a multiplicative factor ($scale$), a translational offset ($offset$), and a clipping threshold ($clip$). During training, these parameters remain fixed ($scale = 1$, $offset = 0$, $clip = \infty$), ensuring the layer does not influence the network's learning. During perturbation testing, these parameters are modified to probe the network's learned features. For each input $x$, the perturbation layer applies the following operation: $y = \min(scale \cdot x + offset, clip)$, where $scale$, $offset$, and $clip$ are adjustable for each unit.

The model was trained on the entire MNIST dataset (rather than using minibatches) for 5000 epochs using Stochastic Gradient Descent (learning rate = 0.001, loss = cross-entropy). Data normalization used $\mu=0.1307$, $\sigma=0.3081$. To ensure statistically significant results, we repeated each experiment 20 times.
\subsection{Perturbation Design}

The core of our experimental design centers on two distinct perturbation types: one targeting distance-based features and the other targeting intensity-based features.

Distance-based features are expected to lie near the decision boundary. By shifting the decision boundary, we increase the distance between active features and the boundary. If these features are critical for classification, this shift should result in reduced model performance. We also seek to maintain the position of intensity features in this operation. For each node, we calculate the output range, scale by the specified percentage, and then apply the offset as a percentage of the range. The perturbation equation for a given percentage $p$ and range $r$ is: $\{scale = (1-p) \cdot r,\, offset = p \cdot r \}$.

We lack a statistical framework for intensity metrics, so we rely on heuristics to identify perturbations that might disrupt them. Two operations are tested: scaling and clipping. Scaling changes the specific value of the intensity feature, while clipping changes the value and removes the ability to distinguish between multiple intensity features.

Scaling simply multiplies node outputs by a scalar value. Distance-features are affected too, but the change is small since they are small. For a scaling percentage $p$, the perturbation equation is $\{scale = p\}$.

Clipping caps activations at threshold value. This destroys information about relative differences among high-activation features. For a cutoff percentage $p$ and range $r$, the equation is $\{clip = p \cdot r\}$

\subsection{Evaluation}

Perturbation ranges were selected to span a broad spectrum to ensure comprehensive evaluation. The ranges overlap to facilitate direct comparison between distance and intensity metrics. All percentages are applied to individual node ranges over the input set. Intensity and cutoff range over $[1\%..1000\%]$. Offset ranges over $[-200\%..100\%]$.

We select a percentage in the perturbation range, calculate and apply $scale$, $offset$ and $clip$ for the active test, evaluate on the entire training set, and calculate the resulting accuracy. We use the training set, and not the test set, to observe how perturbations affect the features learned during training. Changes in accuracy indicate reliance on the perturbed feature type, while stable accuracy suggests that the features are not critical to the model's decisions. The use of the training set ensures a comprehensive assessment with a sufficient number of data points.

%% file: results.tex
\section{Results}

\begin{figure}[ht]
 \centering
 \includegraphics[width=\textwidth]{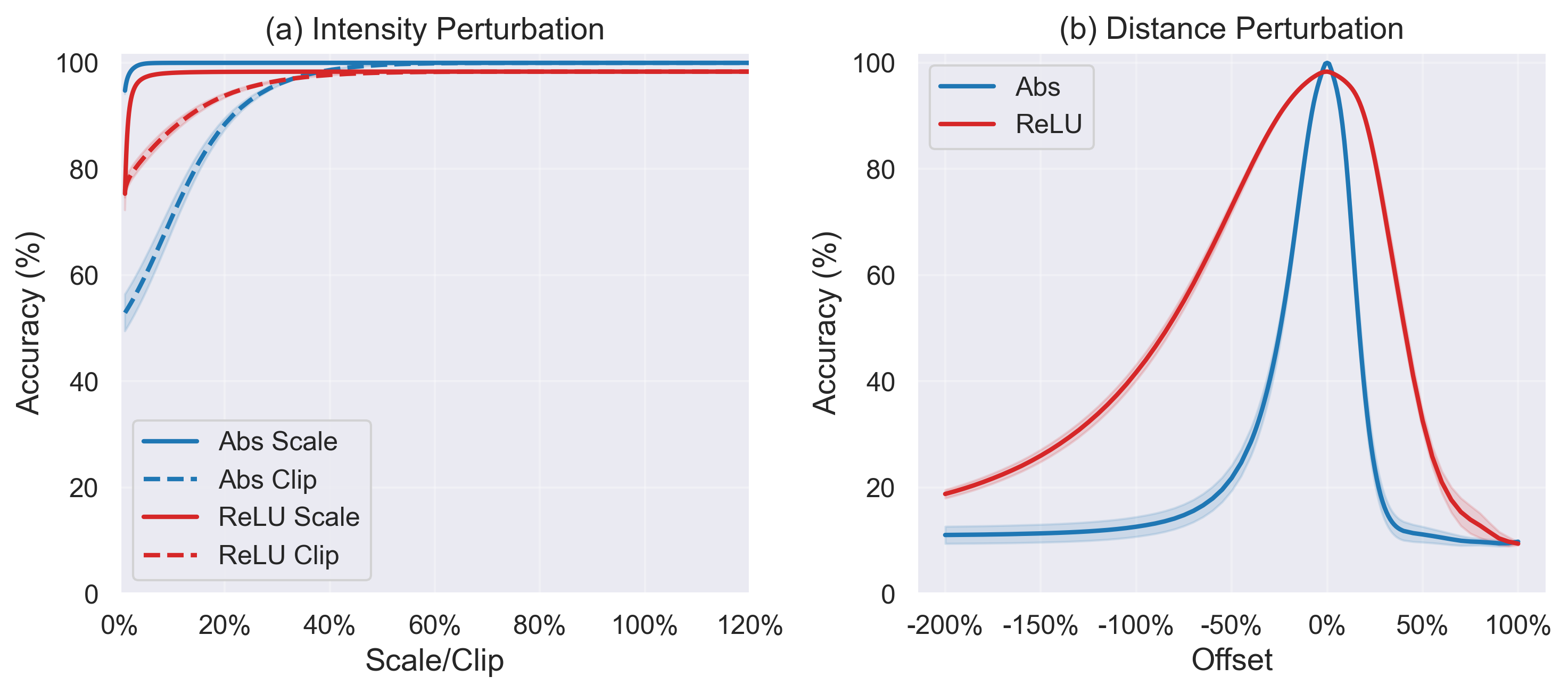}
 \caption{Effects of intensity scaling and distance offset perturbations on model accuracy. Shaded regions represent 95\% confidence intervals across 20 runs.}
 \label{fig:perturbation_analysis}
\end{figure}

Our experiments provide strong empirical support for the theory that the tested models (Abs and ReLU) primarily utilize \emph{distance metrics}, rather than \emph{intensity metrics}, for classification. This means that the models rely on features residing near the decision boundaries for classification, rather than in regions with high activation magnitudes. As shown in Table~\ref{tab:stat_baseline}, both models achieved high accuracy on MNIST before perturbation testing \citep{lecun1998gradient}.

Consistent with theory, both models resist intensity perturbations but are sensitive to distance ones (Figure~\ref{fig:perturbation_analysis}). Specifically, both models maintain their baseline accuracy (approximately 98\% for ReLU and 99\% for Abs) across a wide range of intensity scaling (from 10\% to 200\% of the original output range) and threshold clipping (from 50\% of the maximum activation and above). The minor fluctuations in accuracy observed within these ranges were small and not statistically significant ($p > 0.05$), as detailed in Table~\ref{tab:stat_scale} and Table~\ref{tab:stat_cutoff}. This robustness to intensity perturbations suggests that the models are not heavily reliant on the absolute magnitude of activations, or \emph{intensity metrics}, for classification. This aligns with findings in adversarial example literature, where imperceptible perturbations can drastically alter model predictions \citep{szegedy2013intriguing,goodfellow2014explaining}.

In contrast, both models exhibit a rapid decline in accuracy with relatively small distance offset perturbations. ReLU maintains its baseline accuracy over an offset range from -3\% to +2\% of the activation range, while the Abs model is even more sensitive, falling below 99\% accuracy outside of -1\% to +1\%. These findings, presented in detail in Table~\ref{tab:stat_offset}, underscore the importance of distance metrics, particularly the distances to decision boundaries, in the learned representations for accurate classification.

The high $p$-values associated with the intensity perturbations (see Appendix~\ref{appendix:statistic_tables}) further support our hypothesis. These non-significant results indicate that the observed variations in accuracy under intensity changes are likely attributable to random fluctuations rather than a systematic effect of the perturbations. This reinforces the notion that the models prioritize distance metrics over intensity metrics, focusing on the features close to decision boundaries for classification.

%% file: discussion.tex
\section{Discussion}

\begin{figure}[ht]
    \centering

    \begin{subfigure}[b]{0.49\textwidth}
    \centering
    \includegraphics[width=\textwidth]{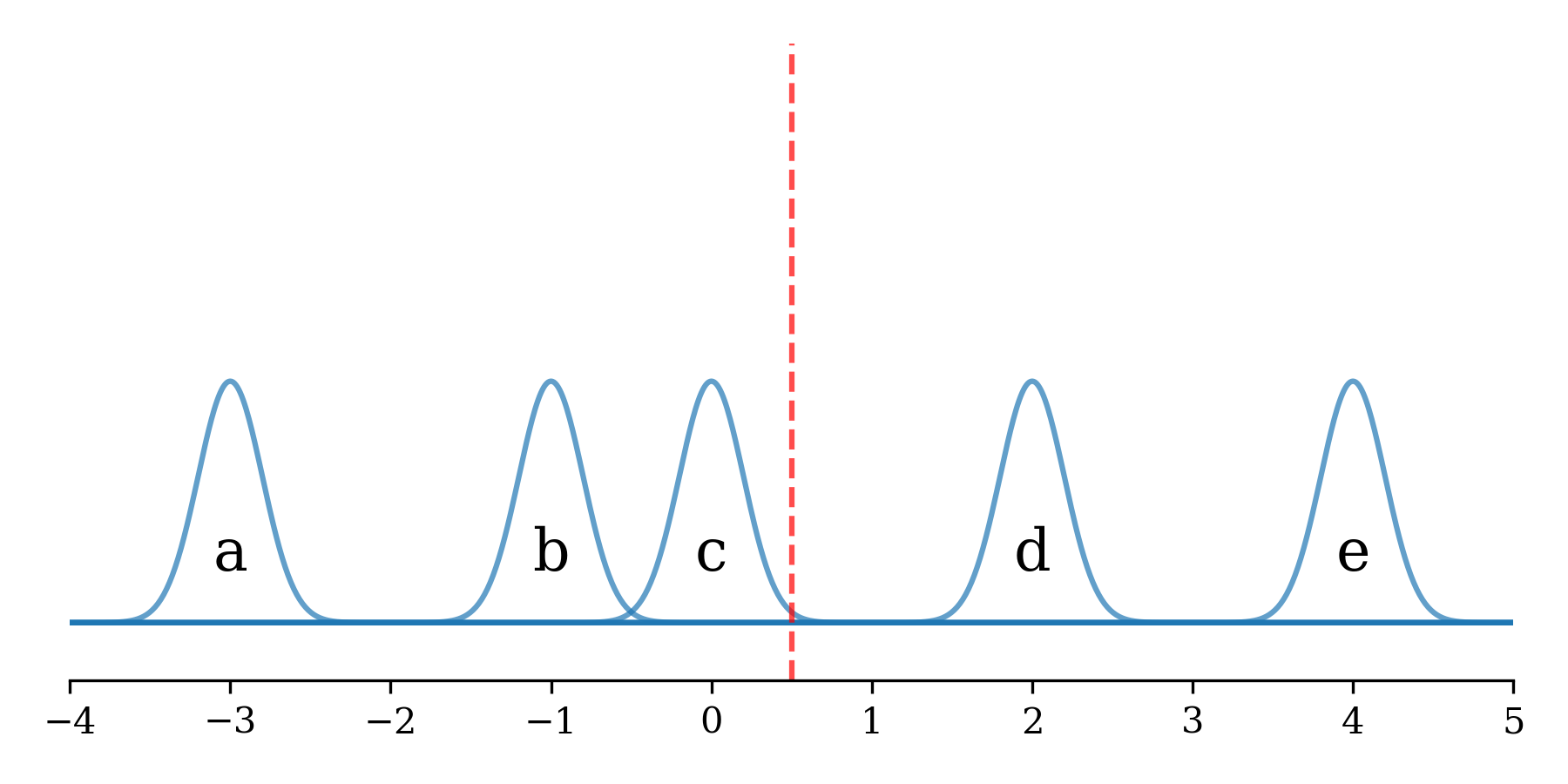}
    \caption{ReLU pre-activation projection}
    \label{fig:relu_pre}
    \end{subfigure}
    \hfill
    \begin{subfigure}[b]{0.49\textwidth}
    \centering
    \includegraphics[width=\textwidth]{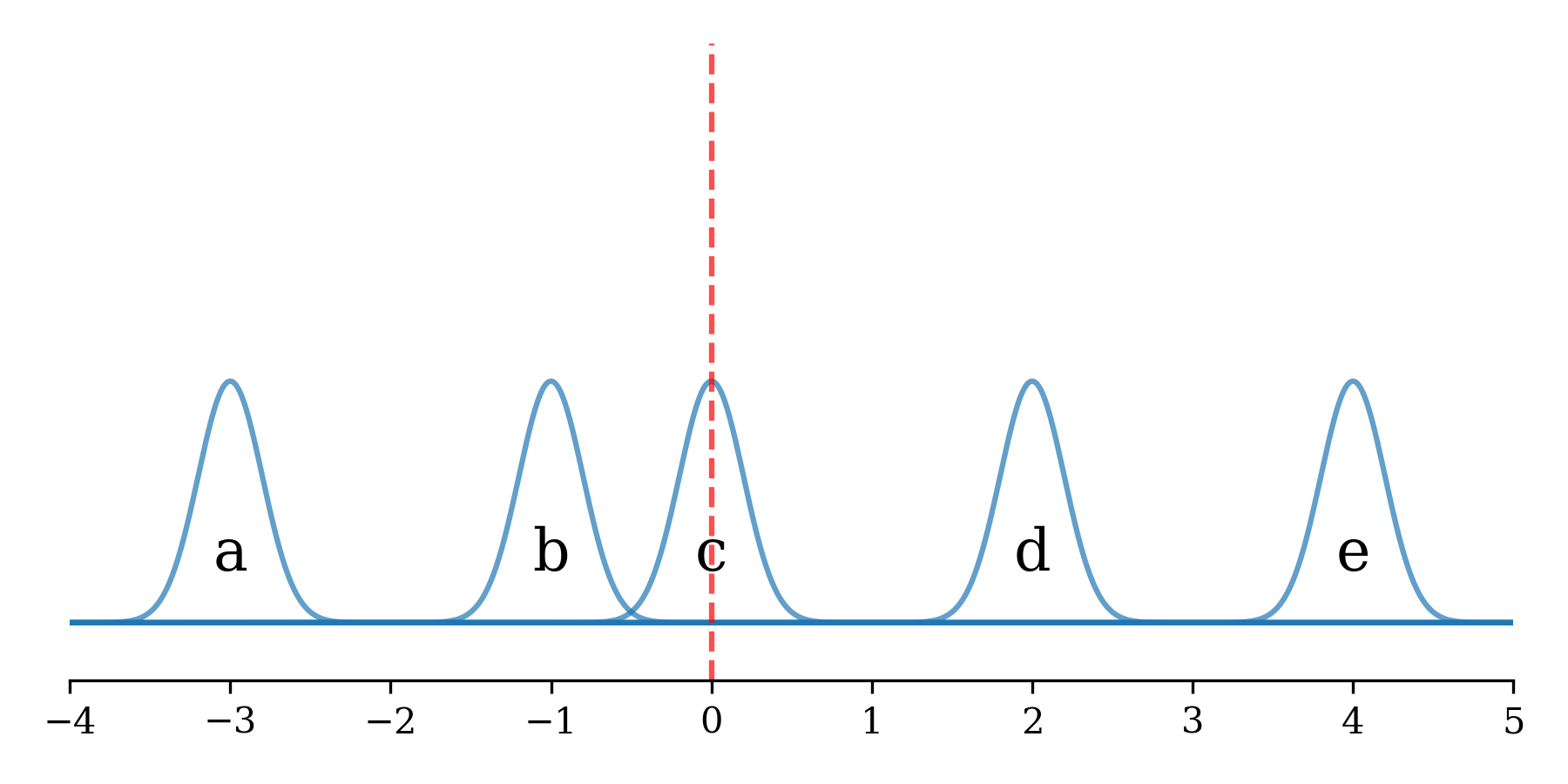}
    \caption{Abs pre-activation projection}
    \label{fig:abs_pre}
    \end{subfigure}

    \begin{subfigure}[b]{0.49\textwidth}
        \centering
        \includegraphics[width=\textwidth]{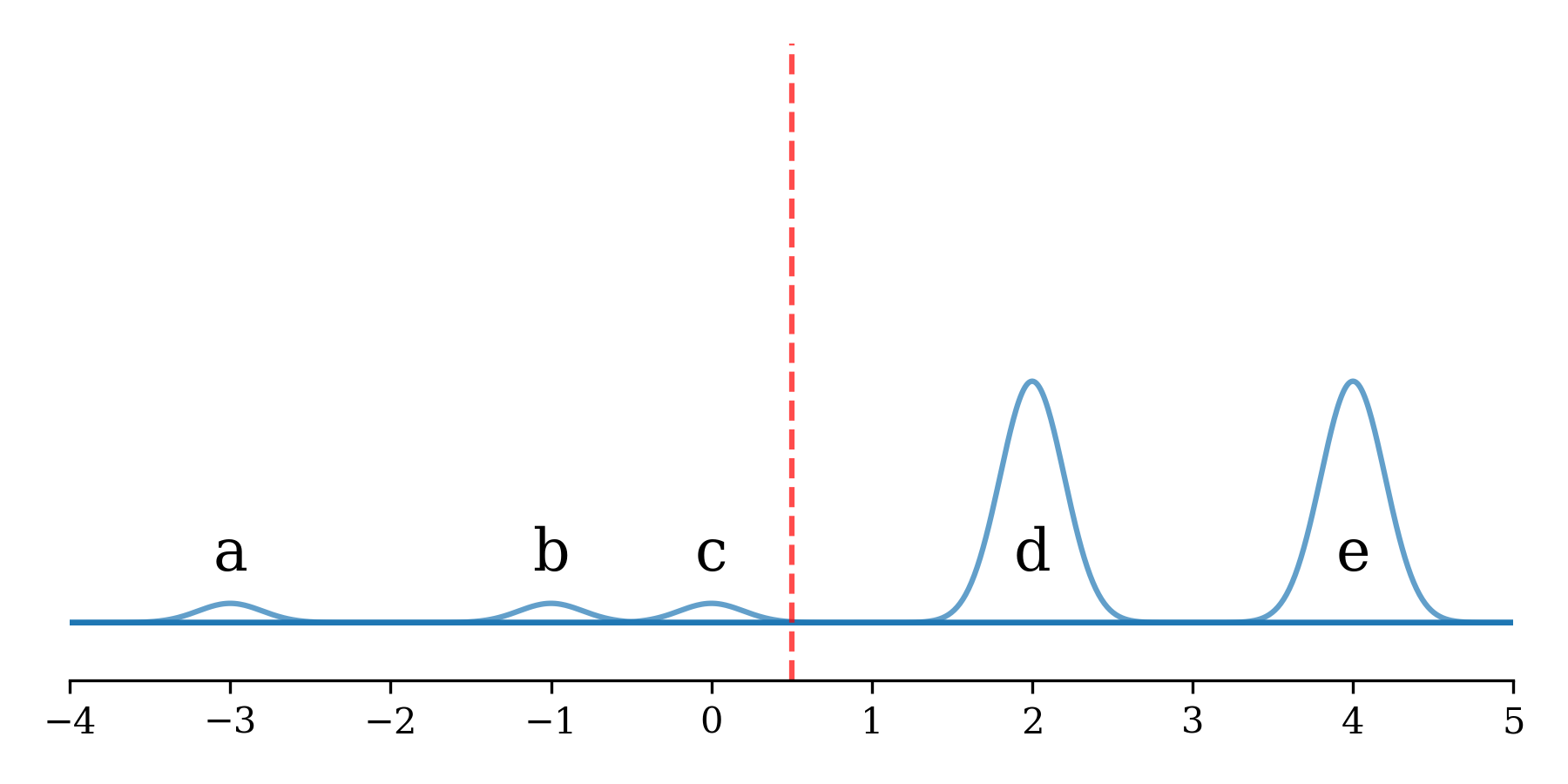}
        \caption{ReLU post-activation response}
        \label{fig:relu_post}
    \end{subfigure}
    \hfill
    \begin{subfigure}[b]{0.49\textwidth}
        \centering
        \includegraphics[width=\textwidth]{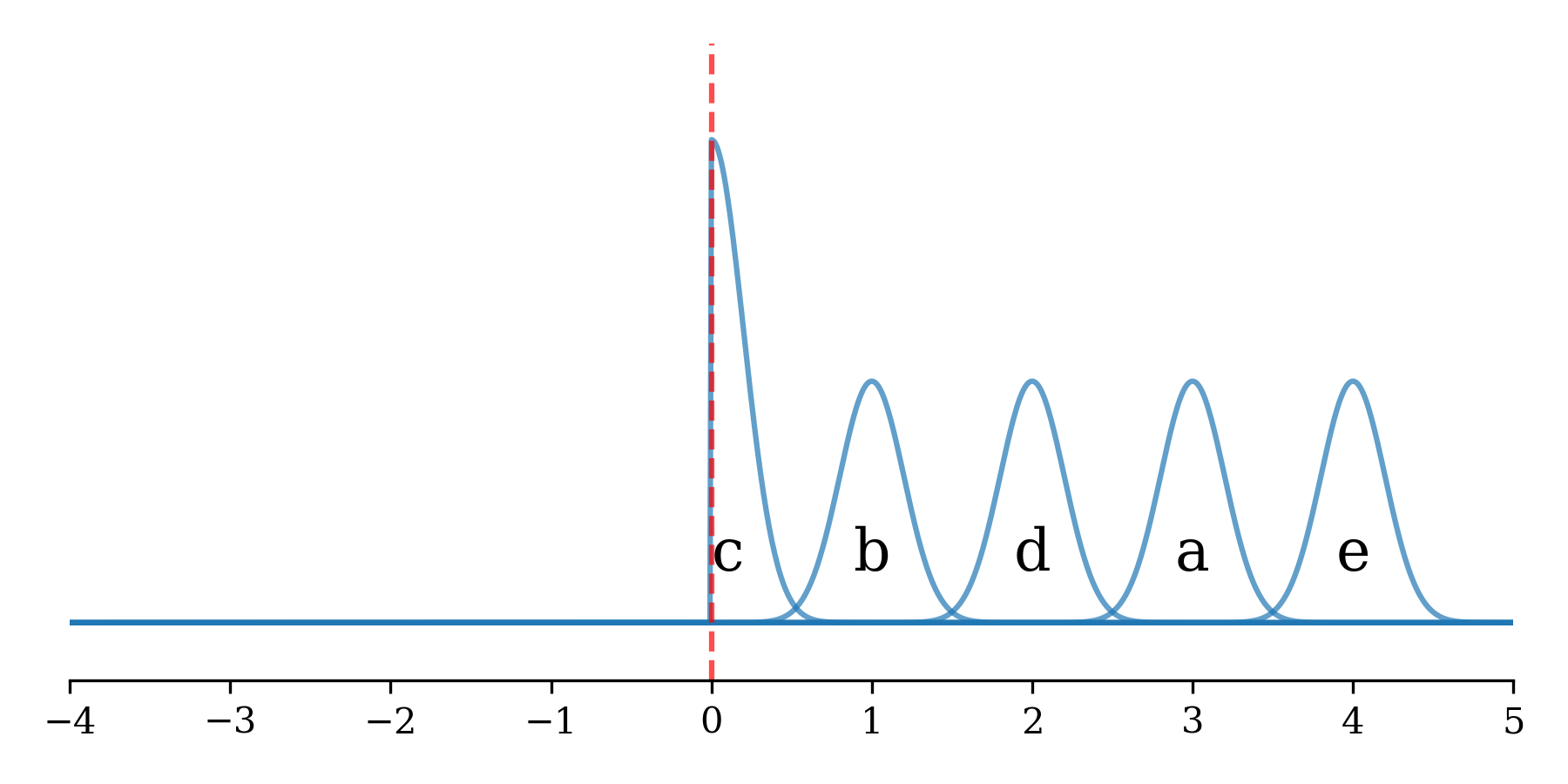}
        \caption{Abs post-activation response}
        \label{fig:abs_post}
    \end{subfigure}

    \caption{This series of figures illustrates how linear nodes process features using ReLU and Absolute Value activation functions. Each blue peak represents a feature (a-e), with the red dashed line showing the decision boundary. The top row shows features after linear projection but before activation. The bottom row shows how ReLU and Absolute Value functions transform these projections, highlighting their distinct effects on feature space.}
    \label{fig:activation_demo}
\end{figure}

We explore how ReLU and Abs activations represent features within a distance metric interpretation. Figure~\ref{fig:activation_demo} illustrates the key differences in how these activation functions process information. In the pre-activation space (Figures~\ref{fig:relu_pre} and~\ref{fig:abs_pre}), both models can learn similar linear projections of input features. ReLU is driven to minimize the active feature $\{c\}$ and ends up being positioned on the positive edge of the distribution. Abs positions the decision boundary through the mean, or possibly the median, of the data. After activation, ReLU sets all features on its dark side to the minimum possible distance: zero. Abs folds the space, moving all distributions on the negative side to the positive side. The ReLU activated node selects for features $\{a,b,c\}$. The folding operation of the Abs activated feature results in $\{c\}$ being the sole feature with the smallest activation value.

\subsection{Offset Perturbations}

\begin{figure}[ht]
    \centering

    \begin{subfigure}[b]{0.49\textwidth}
        \centering
        \includegraphics[width=\textwidth]{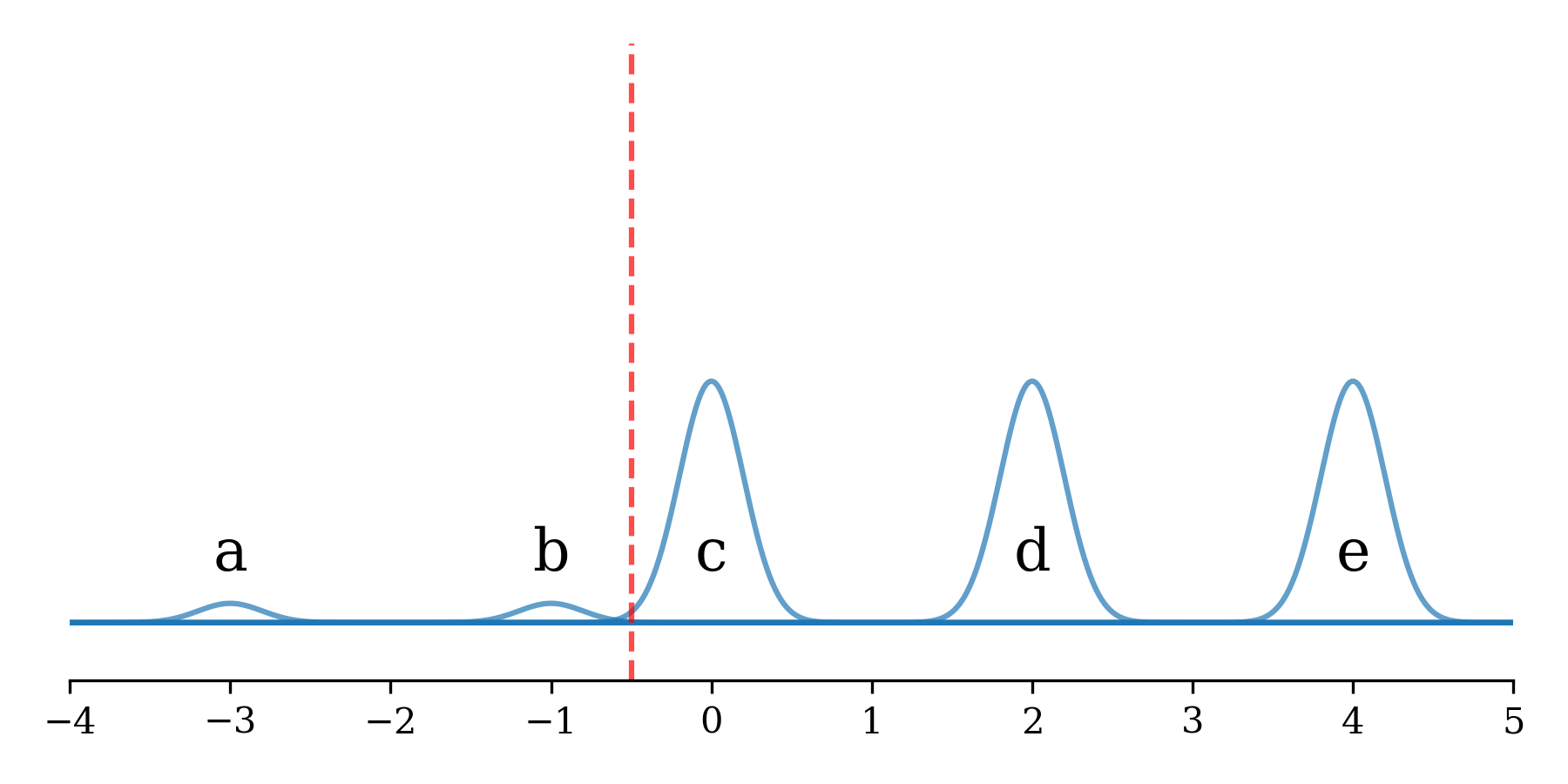}
        \caption{ReLU Negative Offset}
        \label{fig:relu_offset_down}
    \end{subfigure}
    \hfill
    \begin{subfigure}[b]{0.49\textwidth}
        \centering
        \includegraphics[width=\textwidth]{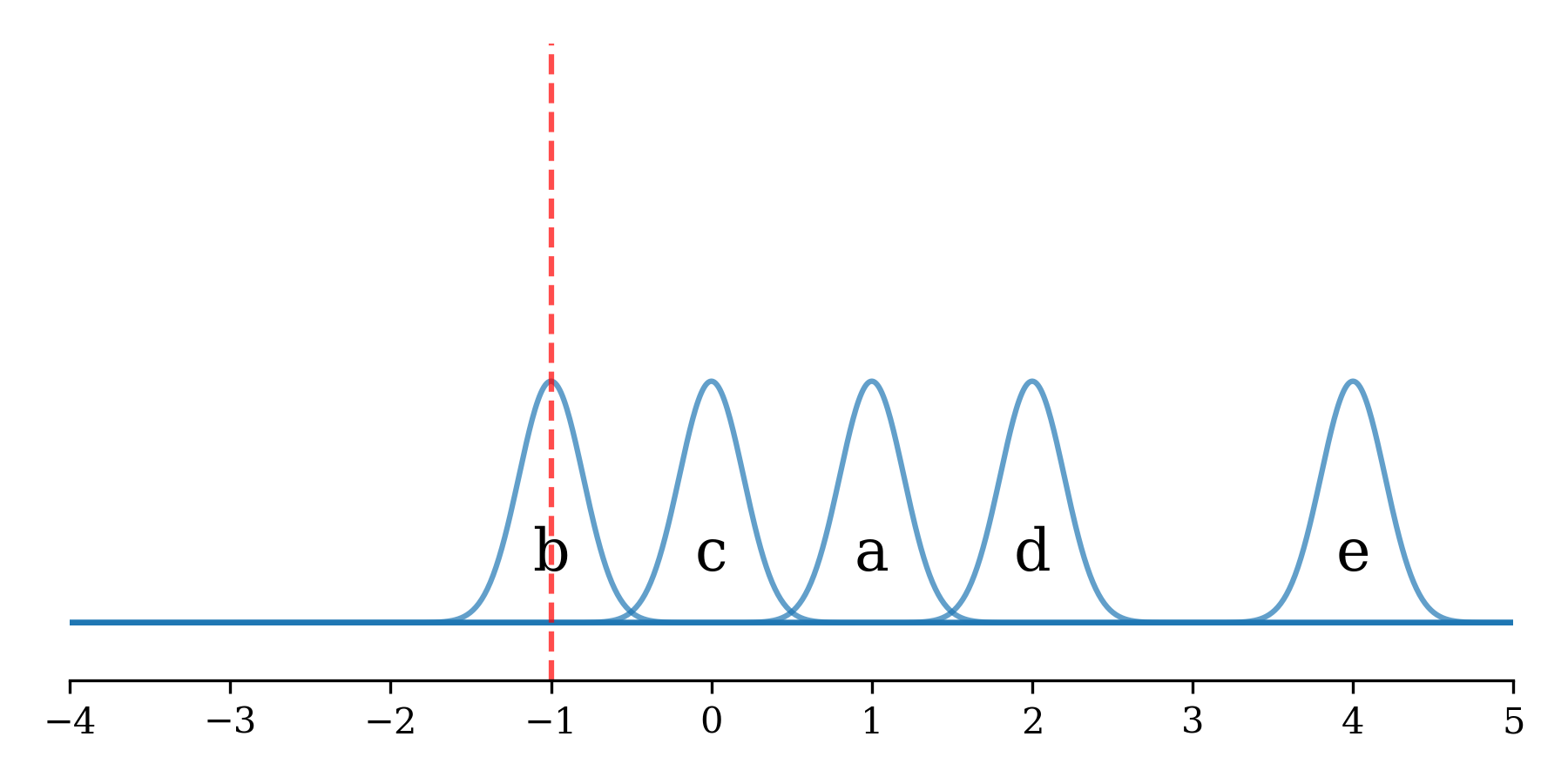}
        \caption{Abs Negative Offset}
        \label{fig:abs_offset_down}
    \end{subfigure}

    \begin{subfigure}[b]{0.49\textwidth}
        \centering
        \includegraphics[width=\textwidth]{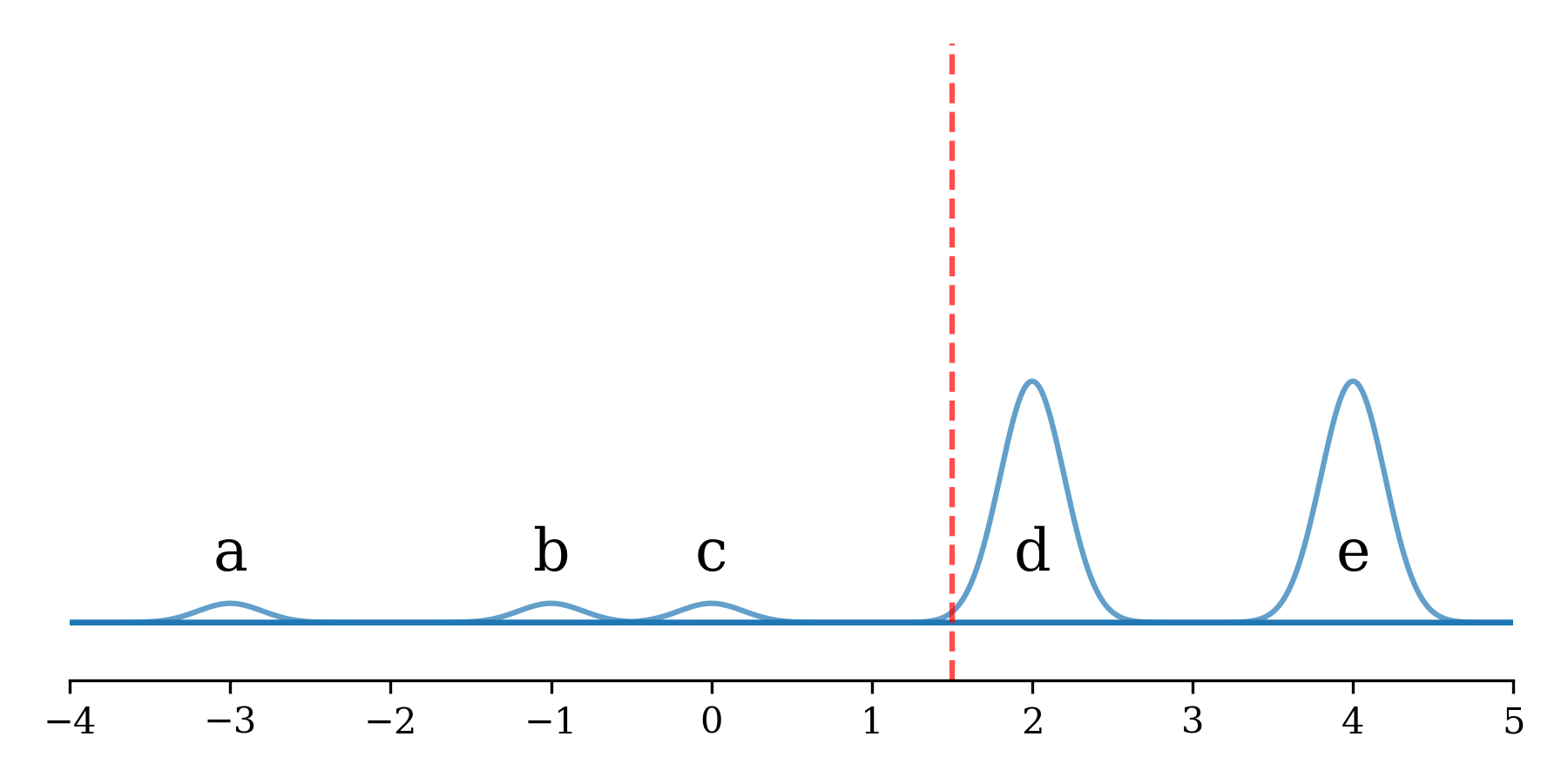}
        \caption{ReLU Positive Offset}
        \label{fig:relu_offset_up}
    \end{subfigure}
    \hfill
    \begin{subfigure}[b]{0.49\textwidth}
        \centering
        \includegraphics[width=\textwidth]{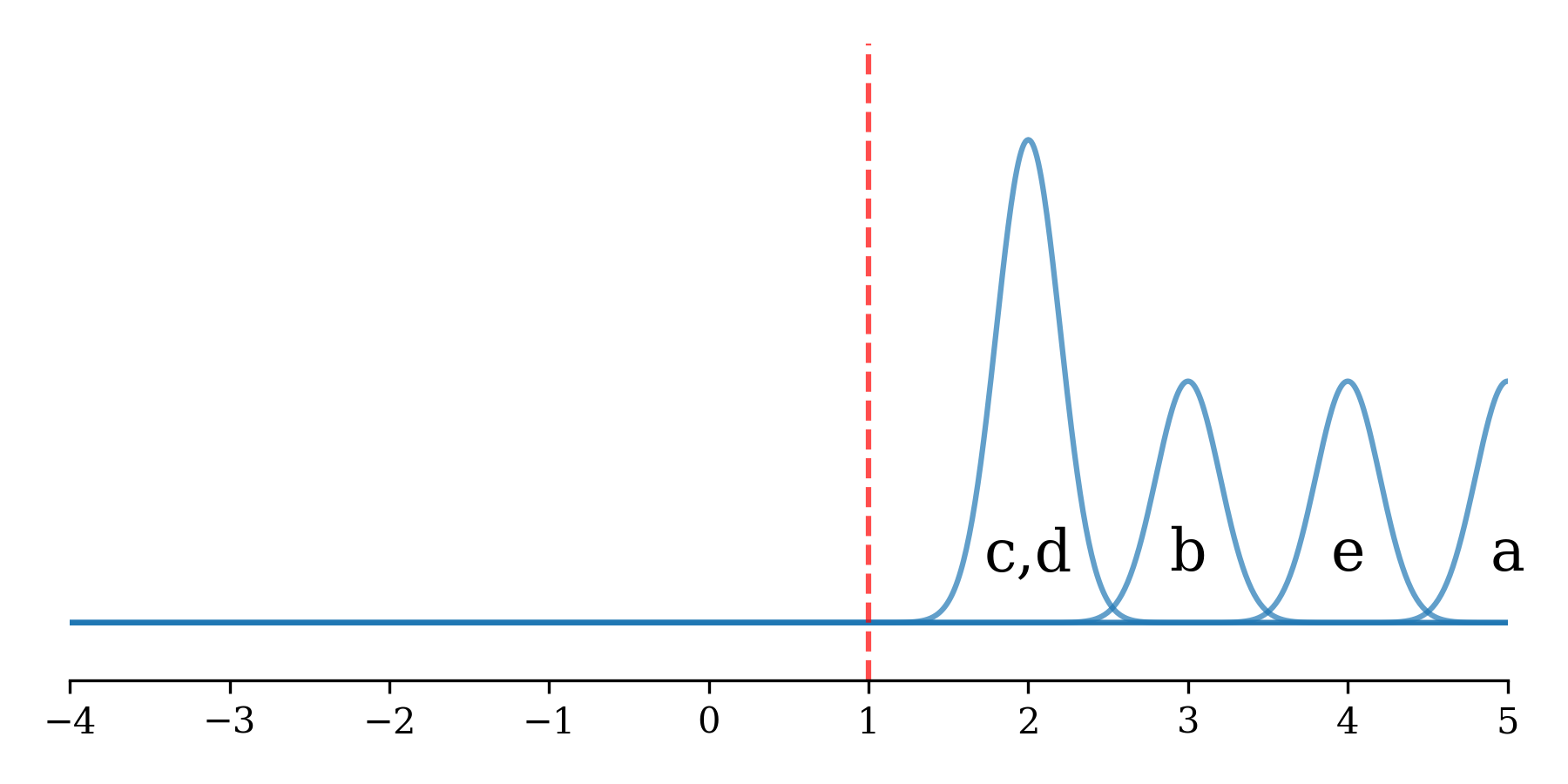}
        \caption{Abs Positive Offset}
        \label{fig:abs_offset_up}
    \end{subfigure}

    \caption{Effects of decision boundary offsets on feature representation. Negative offsets (top row) and positive offsets (bottom row) demonstrate how shifting the decision boundary affects feature selection in ReLU and Abs activated nodes.}
    \label{fig:offset_demo}
\end{figure}

Figure~\ref{fig:offset_demo} illustrates how offset perturbations affect feature selection for both ReLU and absolute value activation functions. With ReLU, offset perturbations modify the set of accepted features. The negative offset (Figure~\ref{fig:offset_demo}\subref{fig:relu_offset_down}) removes feature $c$ from the accepted set, leaving only $\{a,b\}$, while, with this distribution, a positive offset doesn't result in a change (Figure~\ref{fig:offset_demo}\subref{fig:relu_offset_up}). 

In contrast, absolute value activation selects a single feature. A negative offset shifts the selection to feature $b$ instead of the originally trained feature $c$ (Figure~\ref{fig:offset_demo}\subref{fig:abs_offset_down}). A positive offset results in features $\{c,d\}$ having the minimum values (Figure~\ref{fig:offset_demo}\subref{fig:abs_offset_up}). However, neither of these aligns with the decision boundary, effectively resulting in an empty set.

These complete shifts in feature selection for the absolute value activation, compared to the incremental changes with ReLU, explain the more dramatic performance impact observed in Figure~\ref{fig:perturbation_analysis}, even with small offset perturbations.

\subsection{Scale Perturbations}

Analyzing the impact of scaling on intensity features presents a challenge due to the lack of a precise definition for what constitutes an intensity feature. Our experiments demonstrated that scaling activations, which directly modifies their magnitude, did not significantly affect the performance of either ReLU or absolute value networks. This invariance is unexpected if we assume that precise large activations values indicate feature presence.

One possible explanation for this invariance could be the normalization effect of the LogSoftmax operation within the cross-entropy loss function \citep{bridle1990probabilistic}. By renormalizing the output values, LogSoftmax might mitigate the impact of scaling on the relative differences between activations, potentially masking any effects on intensity-based features. However, this does not explain the performance drop observed when activations are scaled down to the magnitudes associated with distance features, suggesting a complex interplay between scaling and the different types of learned features.

\subsection{Cutoff Perturbations}

To further investigate intensity features, we introduced a threshold cutoff perturbation. This perturbation directly targets the ability to distinguish between features based on activation magnitude by clipping activations at a certain threshold. Our results showed a minor performance degradation for cutoff thresholds up to the 50th percentile, followed by a more moderate degradation as the threshold is further reduced. This suggests that the ability to distinguish between features with very high activations might not be critical for classification, especially if subsequent layers utilize sets of features rather than relying on individual activations, as indicated by our analysis of ReLU networks.

While the results of our intensity perturbation experiments generally support our hypothesis that neural networks prioritize distance-based features, the evidence is not as conclusive as with the distance perturbation experiments. Further investigation is needed to fully understand the role of intensity features and their interaction with different activation functions and network architectures.

\subsection{The Problem with Intensity}

Our perturbation analysis appears to support distance-based feature interpretation, but we must address a significant challenge: we cannot definitively disprove intensity-based interpretations due to the lack of a widely accepted definition of what constitutes an intensity feature. This ambiguity has persisted despite decades of research, with various interpretations proposed but no consensus reached. Some studies suggest that intensity features are indicated by maximum activation values, as seen in the foundational work on artificial neurons and perceptrons \citep{mcculloch1943logical,rosenblatt1958perceptron}. Others propose that intensity features might be defined by activation values falling within a specific range, aligning with the concept of confidence intervals or thresholds.

The absence of a clear mathematical foundation for intensity metrics further complicates the matter. Distance metrics like Euclidean and Mahalanobis distances have well-defined statistical measures with clear linear formulations \citep{deza2009encyclopedia,mahalanobis1936generalized}. However, we find no equivalent statistical measure for intensity that can be expressed through a linear equation. This lack of a concrete mathematical basis makes it challenging to design experiments that definitively target and assess intensity features.

Our scaling experiments highlight this difficulty. One might expect that doubling a strong signal (high activation) should make it stronger, yet our networks maintain consistent behavior under scaling. If we propose that relative values between nodes preserve intensity information, this begins to sound suspiciously like a distance metric.

The distance features in the network are easily explained as a Mahalanobis distance of a principal component as described in \citep{oursland2024interpreting}. But what is the statistical meaning behind the intensity features? It implies a complement to the principal component, a principal disponent consisting of an antivector, antivalue, and an unmean. I don't think that principal disponents are real. What looks like an intensity metric is really a distance metric that matches everything except the large value. Perhaps statistical network interpretation has stymied researchers because we have been looking for the mathematical equivalent of Bigfoot or the Loch Ness Monster.

%% file: conclusion.tex
\section{Conclusion}

This paper provides empirical validation for the theoretical connection between neural networks and Mahalanobis distance proposed in \citep{oursland2024interpreting}. Through systematic perturbation analysis, we demonstrated that neural networks with different activation functions implement distinct forms of distance-based computation, offering new insights into their learning and decision-making processes.

Our experiments show that both architectures are sensitive to distance perturbations but resistant to intensity perturbations. This supports the idea that neural networks learn through distance-based representations. The Abs network's performance degrades more dramatically with small offsets than the ReLU network's performance. This may be because the Abs network relies on precise distance measurements, while the ReLU network uses a multi-feature approach.

Both architectures maintain consistent performance under scaling perturbations, which appears to support distance-based rather than intensity-based computation. However, the lack of a precise mathematical definition for intensity metrics makes it difficult to definitively rule out intensity-based interpretations. This limitation highlights a broader challenge in the field: we cannot fully disprove a concept that lacks rigorous mathematical formulation.

These results provide empirical support for the theory that linear nodes naturally learn to generate distance metrics. However, more work is needed to strengthen this theoretical framework, particularly in understanding how these distance computations compose through deeper networks and interact across multiple layers. The evidence presented here suggests that distance metrics may provide a more fruitful framework for understanding and interpreting neural networks than traditional intensity-based interpretations.

%% file: appendix_statistics.tex
\section{Statistic Tables}
\label{appendix:statistic_tables} 

\subsection{Baseline Performance}

The following table presents the detailed performance metrics for both architectures across 20 training runs:

\begin{table}[H]
\centering
\begin{tabular}{lrrr}
\hline
Model & Training Acc (\%) & Test Acc (\%) & Loss \\
\hline
Abs & 99.99 $\pm$ 0.00 & 95.29 $\pm$ 0.20 & 0.0047 $\pm$ 0.0005 \\
ReLU & 98.33 $\pm$ 0.15 & 95.61 $\pm$ 0.14 & 0.0610 $\pm$ 0.0044 \\
\hline
\end{tabular}
\caption{Baseline model performance averaged across 20 training runs (mean $\pm$ standard deviation).}
\label{tab:stat_baseline}
\end{table}

\subsection{Intensity Scale Perturbation Results}

Results for intensity scaling perturbations compared to baseline performance:

\begin{table}[H]
\centering
\begin{tabular}{r|rrr|rrr}
Scale & \multicolumn{3}{|c|}{Abs} & \multicolumn{3}{|c}{ReLU} \\
Change & Acc (\%) & T-stat & P-value & Acc (\%) & T-stat & P-value \\
\hline
1\% & 94.76 & 16.6 & 9.0e-13 & 75.33 & 15.6 & 2.8e-12 \\
5\% & 99.86 & 6.4 & 4.0e-06 & 97.33 & 11.9 & 3.1e-10 \\
10\% & 99.98 & 5.5 & 2.9e-05 & 98.10 & 10.4 & 2.9e-09 \\
25\% & 99.98 & 1.1 & 3.0e-01 & 98.31 & 4.7 & 1.7e-04 \\
50\% & 99.99 & -2.0 & 5.6e-02 & 98.33 & -3.7 & 1.6e-03 \\
Baseline & 99.99 & -1.3 & 2.0e-01 & 98.33 & -3.6 & 1.7e-03 \\
1000\% & 99.99 & -1.6 & 1.3e-01 & 98.33 & -3.2 & 4.9e-03 \\
\end{tabular}
\caption{Effects of intensity scaling on model accuracy. Scale values are shown as percentages of the original range.}
\label{tab:stat_scale}
\end{table}

\subsection{Intensity Cutoff Results}

Results for intensity cutoff perturbations compared to baseline performance:

\begin{table}[H]
\centering
\begin{tabular}{r|rrr|rrr}
Cutoff & \multicolumn{3}{|c|}{Abs} & \multicolumn{3}{|c}{ReLU} \\
Change & Acc (\%) & T-stat & P-value & Acc (\%) & T-stat & P-value \\
\hline
1\% & 52.90 & 28.1 & 6.1e-17 & 75.93 & 36.6 & 4.5e-19 \\
5\% & 60.13 & 24.8 & 6.4e-16 & 82.57 & 27.8 & 7.2e-17 \\
10\% & 71.05 & 21.9 & 6.0e-15 & 87.56 & 23.0 & 2.5e-15 \\
20\% & 88.55 & 21.0 & 1.3e-14 & 93.84 & 20.6 & 1.9e-14 \\
30\% & 96.11 & 24.5 & 7.8e-16 & 96.67 & 18.6 & 1.2e-13 \\
40\% & 98.70 & -25.3 & 4.3e-16 & 97.74 & -23.4 & 1.8e-15 \\
50\% & 99.60 & -26.7 & 1.6e-16 & 98.14 & -26.8 & 1.5e-16 \\
75\% & 99.98 & -26.5 & 1.8e-16 & 98.33 & -27.8 & 7.4e-17 \\
Baseline & 99.99 & -26.5 & 1.8e-16 & 98.33 & -28.1 & 6.3e-17 \\
\end{tabular}
\caption{Effects of intensity cutoff on model accuracy. Cutoff values are shown as percentages of the maximum activation.}
\label{tab:stat_cutoff}
\end{table}

\subsection{Distance Offset Perturbation Results}

Results for distance perturbations (offset) compared to baseline performance:

\begin{table}[H]
\centering
\begin{tabular}{r|rrr|rrr}
Offset & \multicolumn{3}{|c|}{Abs} & \multicolumn{3}{|c}{ReLU} \\
Change & Acc (\%) & T-stat & P-value & Acc (\%) & T-stat & P-value \\
\hline
-200\% & 11.04 & 49.6 & 1.4e-21 & 18.77 & 211.3 & 1.7e-33 \\
-100\% & 12.60 & 48.8 & 1.9e-21 & 41.68 & 78.9 & 2.2e-25 \\
-75\% & 14.79 & 47.1 & 3.9e-21 & 55.16 & 60.2 & 3.8e-23 \\
-50\% & 21.78 & 43.7 & 1.6e-20 & 72.90 & 43.4 & 1.8e-20 \\
-25\% & 49.03 & 32.0 & 5.3e-18 & 90.24 & 33.1 & 2.9e-18 \\
-10\% & 86.25 & -31.5 & 7.3e-18 & 96.55 & -36.7 & 4.1e-19 \\
-5\% & 95.49 & -39.5 & 1.1e-19 & 97.78 & -38.8 & 1.5e-19 \\
-3\% & 97.86 & -41.3 & 4.6e-20 & 98.13 & -39.3 & 1.1e-19 \\
-2\% & 98.95 & -42.1 & 3.2e-20 & 98.24 & -39.4 & 1.1e-19 \\
-1\% & 99.82 & -42.7 & 2.4e-20 & 98.31 & -39.8 & 8.9e-20 \\
Baseline & 99.99 & -42.8 & 2.3e-20 & 98.33 & -40.0 & 8.3e-20 \\
+1\% & 99.81 & -42.7 & 2.4e-20 & 98.26 & -39.6 & 9.8e-20 \\
+2\% & 98.85 & -42.0 & 3.3e-20 & 98.14 & -39.3 & 1.2e-19 \\
+3\% & 97.70 & -41.0 & 5.2e-20 & 97.99 & -38.5 & 1.7e-19 \\
+5\% & 95.00 & -38.3 & 1.8e-19 & 97.62 & -36.6 & 4.5e-19 \\
+10\% & 81.40 & -19.5 & 4.9e-14 & 96.31 & -28.3 & 5.3e-17 \\
+25\% & 23.43 & 15.0 & 5.2e-12 & 81.60 & 14.7 & 8.1e-12 \\
+50\% & 11.14 & 29.7 & 2.2e-17 & 32.54 & 64.3 & 1.1e-23 \\
+75\% & 9.81 & 42.8 & 2.3e-20 & 13.94 & 61.1 & 2.8e-23 \\
+100\% & 9.78 & 44.6 & 1.1e-20 & 9.41 & 493.8 & 1.7e-40 \\
\end{tabular}
\caption{Effects of distance offset on model accuracy. Offset values are shown as percentages of the activation range.}
\label{tab:stat_offset}
\end{table}